\documentclass[conference]{IEEEtran}
\IEEEoverridecommandlockouts
\usepackage{cite}
\usepackage{amsmath,amssymb,amsfonts}
\usepackage{algorithmic}
\usepackage{graphicx}
\usepackage{textcomp}
\usepackage{xcolor}
\def\BibTeX{{\rm B\kern-.05em{\sc i\kern-.025em b}\kern-.08em
    T\kern-.1667em\lower.7ex\hbox{E}\kern-.125emX}}

\usepackage{caption}
\usepackage{tcolorbox}
\usepackage{listings} 
\usepackage{xcolor}   
\usepackage{hyperref}

\tcbset{
    promptbox/.style={
        colframe=blue!50!black,    
        colback=blue!5!white,      
        coltitle=white,           
        fonttitle=\bfseries,      
        arc=2mm,                  
        boxrule=1pt,               
        width=\textwidth,        
        sharp corners=south,      
    },
    codebox/.style={
        colframe=gray!50,          
        boxrule=0.5pt,             
        fontupper=\ttfamily,       
        width=\textwidth,          
    }
}

\begin{document}

\title{Evaluation of Hate Speech Detection using Large Language Models and Geographical
Contextualization}

\author{\IEEEauthorblockN{Anwar Hossain}
\IEEEauthorblockA{\textit{Department of Computer Science} \\
\textit{Iowa State University}\\
Iowa, USA \\
ahzahid@iastate.edu}
\and
\IEEEauthorblockN{Monoshi Roy}
\IEEEauthorblockA{\textit{Department of Computer Science} \\
\textit{Iowa State University}\\
Iowa, USA \\
monoshi@iastate.edu}
\and
\IEEEauthorblockN{Swarna Das}
\IEEEauthorblockA{\textit{Department of Computer Science} \\
\textit{Iowa State University}\\
Iowa, USA \\
swarna@iastate.edu}
}
\maketitle

\begin{abstract}
The proliferation of hate speech on social media is one of the serious issues that is bringing huge impacts to society: an escalation of violence, discrimination, and social fragmentation. The problem of detecting hate speech is intrinsically multifaceted due to cultural, linguistic, and contextual complexities and adversarial manipulations. In this study, we systematically investigate the performance of LLMs on detecting hate speech across multilingual datasets and diverse geographic contexts. Our work presents a new evaluation framework in three dimensions: binary classification of hate speech, geography-aware contextual detection, and robustness to adversarially generated text. Using a dataset of 1,000 comments from five diverse regions, we evaluate three state-of-the-art LLMs: Llama2 (13b), Codellama (7b), and DeepSeekCoder (6.7b). Codellama had the best binary classification recall with 70.6\% and an F1-score of 52.18\%, whereas DeepSeekCoder had the best performance in geographic sensitivity, correctly detecting 63 out of 265 locations. The tests for adversarial robustness also showed significant weaknesses; Llama2 misclassified 62.5\% of manipulated samples. These results bring to light the trade-offs between accuracy, contextual understanding, and robustness in the current versions of LLMs. This work has thus set the stage for developing contextually aware, multilingual hate speech detection systems by underlining key strengths and limitations, therefore offering actionable insights for future research and real-world applications.

\end{abstract}


\section{Introduction}

Hate speech on social media has emerged as a pervasive and harmful phenomenon, with far-reaching societal implications. Defined as language that degrades, intimidates, or dehumanizes individuals or groups based on their identity, race, gender, disability, sexual orientation, or other characteristics, hate speech contributes to the escalation of violence, discrimination, and societal polarization \cite{ndtv_riots, guardian_durga_puja}. Social media platforms have become major vehicles for spreading harmful ideologies, despite their utilities in fostering global connectivity. Recent studies show alarming trends: the European Commission reported a 20\% increase in hate speech on major platforms between 2020 and 2023, while in the United States, the Anti-Defamation League found a 21\% year-over-year increase in online hate incidents. The subcontinent has been especially susceptible to the aftereffects of online hate speech. In India, online propaganda has led to numerous communal riots, killing more than 1600 people between 2004 and 2017 \cite{ndtv_riots}. In Bangladesh, hate speech took its course during the Durga Puja riots in 2021, killing seven and injuring over 150 people \cite{guardian_durga_puja}. These tragedies further pinpoint the dire need for scalable, reliable, and context-sensitive mechanisms for detecting hate speech.

Detecting hate speech is intrinsically a very challenging task. While existing Machine Learning based methods work relatively well for explicit hate speech detection, most fail in the case of subtle or implicit hate speech manifestations as reported in related works \cite{related_work1}. In particular, sarcastic or coded languages mostly bypass traditional systems. Additionally, the linguistic and cultural diversity in regions like South Asia exacerbates the challenge, as hate speech varies significantly across languages, dialects, and social contexts \cite{related_work2}. To compound the issue, adversarial attacks, where hateful intent is disguised through paraphrasing or obfuscation, expose the fragility of current detection frameworks \cite{adversarial_text}. These limitations necessitate innovative approaches that integrate contextual understanding, multilingual adaptability, and robustness against adversarial manipulation.

LLMs like GPT-4 and Codellama provide a promising avenue to solve these challenges. Trained on large multilingual datasets, LLMs possess state-of-the-art contextual understanding, semantic analysis, and transfer learning capabilities \cite{llm_advances}. Initial studies have shown their potential in a wide variety of tasks ranging from sentiment analysis to text classification. However, their performance on hate speech detection, specifically on multilingual and geographically diverse datasets, is still relatively unexplored \cite{hate_speech_llm_review}. Our research fills this gap by rigorously evaluating the performance of state-of-the-art LLMs in hate speech detection under three key dimensions. Firstly, we assess the accuracy, precision, recall, and F1-score of LLMs in the binary classification of hate speech in five languages, namely Arabic, Bengali, Hindi, Chinese, and Russian. We also probe into the ability of the models to inculcate geographical and cultural contexts into their classification decisions. At last, we test resilience against adversarially generated text designed to obscure hateful intent. These are achieved by collecting a dataset of 10,000 multilingual comments from publicly available sources such as Kaggle, Hatebase, and community-driven datasets \cite{kaggle_dataset, hatebase_dataset}. Each comment was annotated for hate speech content and geographical origin. Non-English comments were translated into English using Google Translate and LLM-based translation techniques, ensuring sentiment preservation. The dataset is uniquely suited for comprehensive evaluation due to the diversity of cultural and linguistic nuances.

Our approach uses prompt engineering to enhance the performance of LLMs. For hate speech detection, we designed structured prompts with in-context examples, which helped the models understand nuanced text. In adversarial robustness tests, we used GPT-4 to generate synthetic adversarial samples by paraphrasing or injecting noise into original comments. These adversarial samples were used to evaluate the consistency and resilience of LLM predictions. Three advanced LLMs were evaluated: Llama2 (13b), Codellama (7b), and DeepSeekCoder (6.7b). Codellama had the best recall (70.6\%) and F1-score (52.18\%) for hate speech detection but only correctly predicted 10.5\% of locations. DeepSeekCoder was more geographic-sensitive but with a lower classification performance. Adversarial testing showed significant vulnerabilities; Llama2 misclassified 62.5\% of adversarial samples. These results indicate the trade-offs in the current state of LLMs on accuracy, sensitivity, and robustness. Our contributions are threefold:
\begin{itemize}
    \item Showing the importance of geographically contextualized datasets for hate speech detection.
    \item Performing an end-to-end evaluation framework to evaluate LLMs across multilingual, adversarial, and geographic dimensions.
    \item Discussing the limitations and suggesting ways of fine-tuning LLMs for improved contextual understanding and robustness.
\end{itemize}

The rest of the paper is organized as follows: Section II reviews the related work on hate speech detection and LLM applications. Section III describes the dataset and experimental setup. In Sect. IV, it presents the technical approach involving prompt engineering and adversarial sample generation. Section V presents the results of the evaluation and Sect. VI discusses those. Finally, Sect. VII outlines the threats to validity, and Sect. VIII concludes the paper with directions for future research.

\section{Related Work}
In recent years, NLP and large language models have improved hate speech detection. LLMs have been contributing to enhancing both performance and interpretability in hate speech detection. Models employ different techniques like rationale extraction to identify and highlight elements, i.e., derogatory words or contextual reasoning behind classifications \cite{nirmal2024towards}.

Different prompting strategies, e.g., general prompts, definition-based prompts, few-shot learning, and chain-of-thought (CoT) were explored for detecting hate speech. Study shows prompting strategy played a critical role in reasoning and significantly improved contextual understanding and classification accuracy \cite{guo2023investigation}. With these strategies, LLMs can effectively use their extensive pre-trained knowledge, even when fine-tuning is not possible. CoT reasoning, in particular, has been shown to enhance the model’s ability to detect implicit hate speech by breaking down complex tasks into logical steps \cite{yang2023hare}. Zero-shot learning approaches using instruction-tuned LLMs have demonstrated equal or better performance compared to the fine-tuned models in low-resource settings. \cite{plaza2023respectful} It suggests the careful selection of verbalizers and prompts helps the models to work with a wide range of datasets and languages when the labeled data is scarce. 
\cite{yang2023hare} emphasize the importance of explainability techniques in the context of LLM-based hate speech detection. CoT reasoning offers detailed, step-by-step rationales for predictions. By combining explanations generated by machines with annotations made by humans, models are able to achieve better alignment with the requirements that are uniquely associated with the dataset. This hybrid approach shows improved classification accuracy and interpretability.
\cite{roy2023probing}

Recently, researchers have focused on fine-tuning LLMs with smaller parameters, which shows promising results on the English hate speech dataset \cite{sen2024hatetinyllmhatespeech}. Vision-based models have also been used to detect hate speech from images and their captions, keeping the cultural context in mind \cite{bui2024multi3hatemultimodalmultilingualmulticultural}. However, though the current research provides significant insights into hate speech detection, minimal attention has been given to the evaluation of LLMs on hate speech detection, keeping the geographical context in mind. Our study aims to bridge the gap in LLM's hate speech evaluation while incorporating geographical information in our evaluation process. This focus will ensure the future directions and challenges in addressing hate speech detection with LLMs across multiple languages and regions.

\section{Technical Approach}
As we wanted to evaluate the capabilities of LLMs in Hate Speech Detection, the first problem we faced was that most of the LLMs were not trained on the multi-lingual corpus. We aimed to evaluate hate speech across multiple languages like Arabic, Bengali, Hindi, etc. Hence, we decided to translate all the sample comments into English and prompt the LLMs based on the translated English comments. For the language translation process, we used Google's translation API \cite{googlecloudtranslation}. However, the quality of Google translation may not be consistent across all the languages, and our work might be substantially dependent on the quality of the translation process. So, we decided to adopt LLM translation and prompted the models to keep the sentiment of the original text intact. Pre-trained models might not be a good approach for language translation; fine-tuning can be a better approach for this application. However, we haven't used any fine-tuning on the models for this project to improve the sentiment-based translation task. We have used two different prompts for the hate speech detection evaluation and translating the language with LLM. For hate speech evaluation, we used the prompt structure in Fig.1, where we used three different language comments for in-context examples. We used specific tags like [ANSWER] [/ANSWER] to dictate the LLMs to give the prediction in particular tags so that we could quickly parse the responses of the LLMs by just extracting the prediction inside these tags. For translation, we used the prompt in Fig. 2. In this case, we used two in-context examples.\newline

We also analyzed the robustness of LLMs responses by promoting the models with some adversarial examples. For this task, we selected 50 English language hate-speech and non-hate speech samples and prompted LLM (GPT-4) to create adversarial samples of those samples to flip the labels for the LLM predictions. Then we prompted (fig. 3) the LLMs with both original and adversarial samples and analyzed how the models predict with the adversarial samples. 
\begin{figure}[h!]
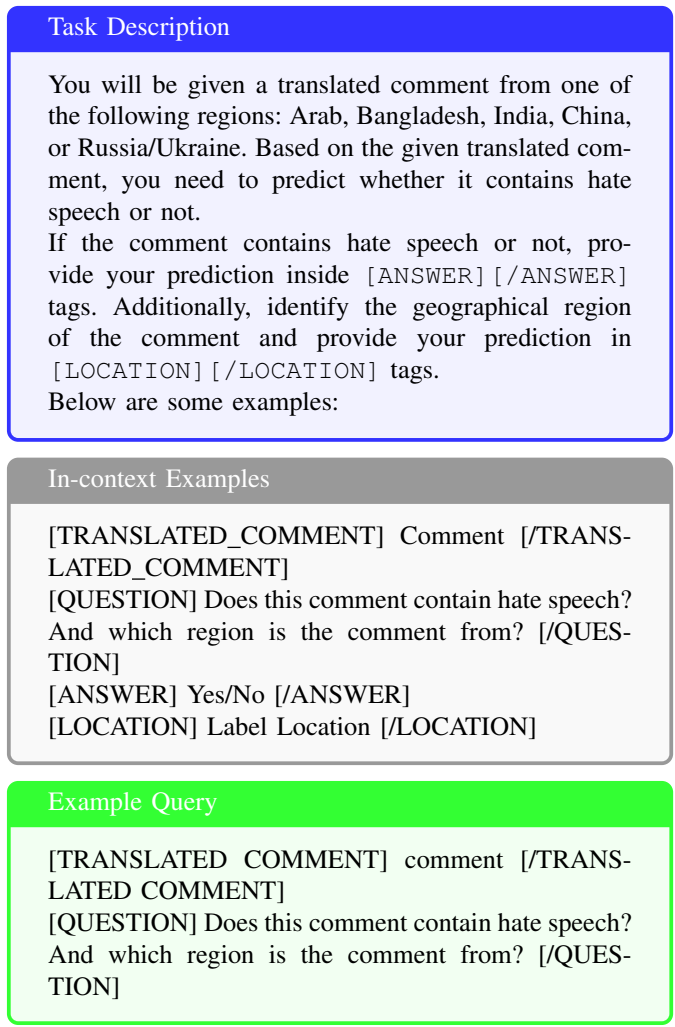

\begin{flushleft}
\begin{tcolorbox}[width=\linewidth, colback=blue!5, colframe=blue!80, title=Task Description]
You will be given a translated comment from one of the following regions: Arab, Bangladesh, India, China, or Russia/Ukraine. Based on the given translated comment, you need to predict whether it contains hate speech or not. 

If the comment contains hate speech or not, provide your prediction inside \texttt{[ANSWER][/ANSWER]} tags. Additionally, identify the geographical region of the comment and provide your prediction in \texttt{[LOCATION][/LOCATION]} tags.

Below are some examples:
\end{tcolorbox}

\begin{tcolorbox}[width=\linewidth, colback=gray!5, colframe=gray!80, title=In-context Examples]
[TRANSLATED\_COMMENT] Comment [/TRANSLATED\_COMMENT] \newline
[QUESTION] Does this comment contain hate speech? And which region is the comment from? [/QUESTION] \newline
[ANSWER] Yes/No [/ANSWER] \newline
[LOCATION] Label Location [/LOCATION]
\end{tcolorbox}

\begin{tcolorbox}[width=\linewidth, colback=green!5, colframe=green!80, title=Example Query]
[TRANSLATED COMMENT] comment [/TRANSLATED COMMENT] \newline
[QUESTION] Does this comment contain hate speech? And which region is the comment from? [/QUESTION]
\end{tcolorbox}
\end{flushleft}
\caption{Prompt Structure for Hate Speech Evaluation.}
\label{fig:task_description}
\end{figure}

\begin{figure}[h!]
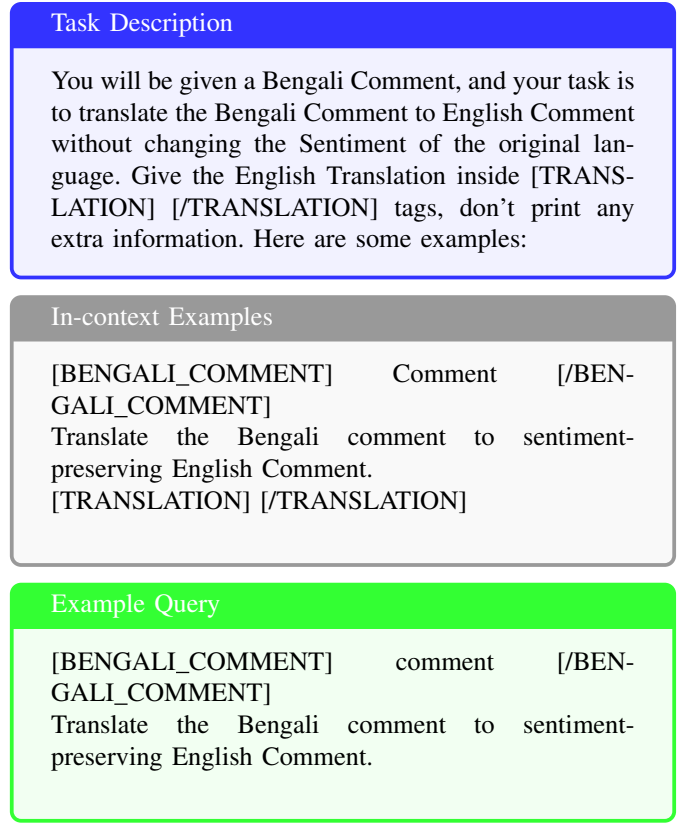

\begin{flushleft}
\begin{tcolorbox}[width=\linewidth, colback=blue!5, colframe=blue!80, title=Task Description]
You will be given a Bengali Comment, and your task is to translate the Bengali Comment to English Comment without changing the Sentiment of the original language. Give the English Translation inside [TRANSLATION] [/TRANSLATION] tags, don't print any extra information.
Here are some examples:
\end{tcolorbox}

\begin{tcolorbox}[width=\linewidth, colback=gray!5, colframe=gray!80, title=In-context Examples]
[BENGALI\_COMMENT] Comment [/BENGALI\_COMMENT] \newline
Translate the Bengali comment to sentiment-preserving English Comment. \newline
[TRANSLATION] [/TRANSLATION] \newline
\end{tcolorbox}

\begin{tcolorbox}[width=\linewidth, colback=green!5, colframe=green!80, title=Example Query]
[BENGALI\_COMMENT] comment [/BENGALI\_COMMENT] \newline
Translate the Bengali comment to sentiment-preserving English Comment. \newline
\end{tcolorbox}
\end{flushleft}
\caption{Prompt Structure for Language Translation.}
\label{fig:task_description}
\end{figure}

\begin{figure}[h!]
\begin{flushleft}
\begin{tcolorbox}[width=\linewidth, colback=blue!5, colframe=blue!80, title=Task Description]
You will be given a comment of English Language. Based on the given comment, you need to predict whether it contains hate speech or not. If the comment contains hate speech, provide your prediction inside [ANSWER][/ANSWER] tags. Below are some examples:
\end{tcolorbox}

\begin{tcolorbox}[width=\linewidth, colback=gray!5, colframe=gray!80, title=In-context Examples]
[COMMENT] Comment [/COMMENT] \newline
[QUESTION] Does this comment contain hate speech?[/QUESTION] \newline
[ANSWER] Yes/No [/ANSWER] \newline
\end{tcolorbox}

\begin{tcolorbox}[width=\linewidth, colback=green!5, colframe=green!80, title=Example Query]
[COMMENT] Comment [/COMMENT] \newline
[QUESTION] Does this comment contain hate speech?[/QUESTION] \newline
\end{tcolorbox}
\end{flushleft}
\caption{Prompt Structure for Language Translation.}
\label{fig:task_description}
\end{figure}

\section{Evaluation}
We can divide our evaluation into some subsections, at first we evaluated three LLMs (llama2 \cite{llama2_13b}, codellama \cite{codellama_7b} and deepseekcoder \cite{deepseek_coder_6_7b}), on our dataset for hate speech detection as a binary classification. After that, we evaluated the capabilities of the models to predict the correct geography of the hate speech, given it predicted the hate speech correctly. Then, we studied how these three models work with the synthetic adversarial examples. And finally, we added the case studies for how LLM translated comments perform and examples of how using adversarial samples misclassify the prediction.

\subsection{Hate Speech Evaluation}
TABLE I shows the result of hate speech evaluation on LLMs. Pre-trained LLMs are not performing significantly well on hate speech detection tasks. Codellama performs well on our three evaluated models in all metrics, substantially surpassing the other two. Codellama correctly predicted the hate speech comments 71\%(Recall: 70.60) of the time. However, as the overall accuracy of Codellama seems to be pretty poor, we can deduce that for most of the comments, Codellama's predictions are biased toward hate speech class. Deepseekcoder ranks second among the three models, with an accuracy of 26.40\% and a recall value of 53\%. Llama2's comparative performance was lower than the other two models in terms of all metrics. Fig. 4. shows the comparative bar-chart of the performance metrics across all three models. 

\begin{table}[h!]
\captionsetup{position=top} 
\caption{Performance comparison of models on key metrics.}
\centering
\begin{tabular}{|l|c|c|c|c|}
\hline
\textbf{Model}          & \textbf{Accuracy (\%)} & \textbf{Precision} & \textbf{Recall} & \textbf{F1 Score} \\ \hline
Llama2 (13b)            & 16.04                 & 23.74              & 33.10           & 27.65             \\ \hline
Codellama (7b)          & 35.30                 & 41.38              & 70.60           & 52.18             \\ \hline
DeepSeek (6.7b)         & 26.50                 & 34.64              & 53.00           & 41.90             \\ \hline
\end{tabular}
\label{tab:model_comparison}
\end{table}

\begin{figure}[h!]
\centering
\includegraphics[width=0.9\linewidth, height=0.3\textheight]{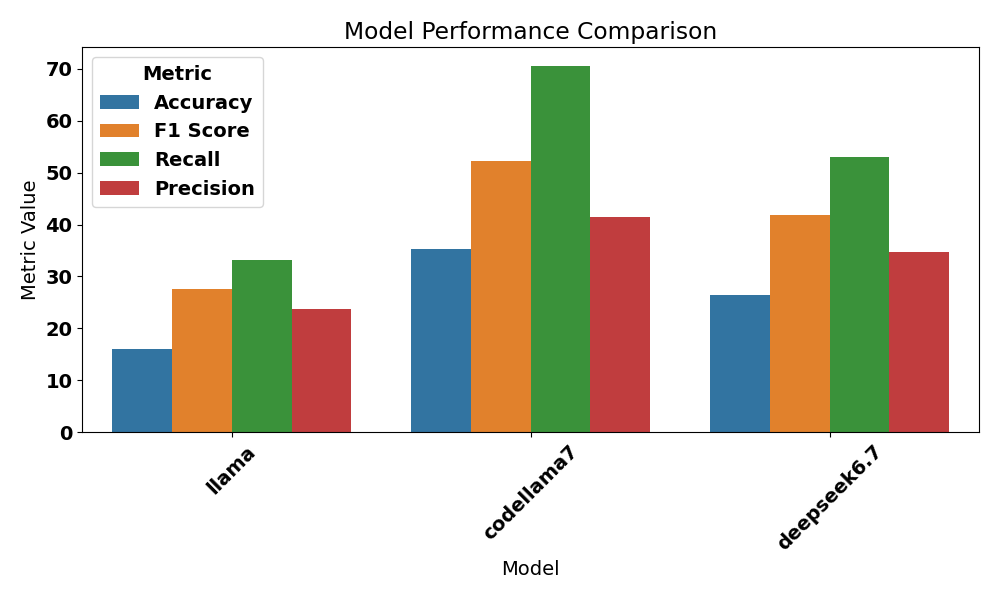} 
\caption{Model Performance on Hate Speech Detection.}
\label{fig:evaluation_process}
\end{figure}

\subsection{Geography Detection Capabilities of the Models}
We also analyzed the capabilities of the LLM models on geographic location. For this, we are only interested in the comments predicted as hate speech by the models. We also analyzed that model's location prediction if any comment is predicted as hate speech. TABLE II shows the geographic prediction capabilities of the models. Interestingly, we observe that, although, Codellama was performing well in hate speech detection, its ability to detect hate speech location is very poor. In total, deepseek correctly predicted 63 locations out of 265 correct predictions; however, in terms of percentage, llama2 is outperforming the other two (37 out of 94). Fig.5 shows the comparative bar chart of this analysis.

\begin{table}[h!]
\captionsetup{position=top} 
\caption{Model Performance Comparison for Correct Hate Speech Prediction and Correct Location}
\centering
\begin{tabular}{|l|c|c|}
\hline
\textbf{Model} & \textbf{Correct HateSpeech Prediction} & \textbf{Correct Location} \\
\hline
Llama2 (13b) & 94  & 37 \\
Codellama (7b) & 353 & 37 \\
deepseek6.7 & 265 & 63 \\
\hline
\end{tabular}
\label{tab:location_comparison}
\end{table}

\begin{figure}[h!]
\centering
\includegraphics[width=0.9\linewidth, height=0.3\textheight]{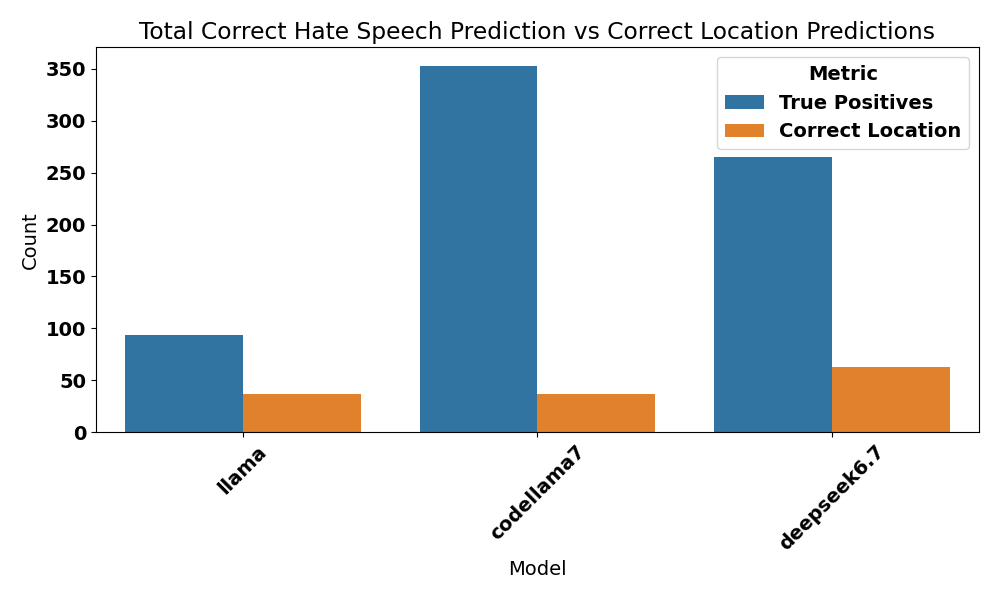} 
\caption{Model Performance on Hate Speech Location.}
\label{fig:evaluation_process}
\end{figure}

\subsection{Evaluation on Adversarial Synthetic Dataset}
We evaluated three LLM models on the synthetic dataset we developed using GPT 4. In this experiment, we focused on whether our synthetic samples can change the label of the predictions of the original samples. Table III depicts the results of this experiment; in the Correct Prediction column, we are showing out of 50 original samples how many samples were correctly predicted by the models. Deepseekcoder outperforms the other two models in predicting the correct English hate speech or benign comments. Half of the comments were predicted correctly by deepseekcoder; on the other hand, llama and codellama correctly predicted 24 and 22 samples. The third column, Adversarial Misclassify, denotes how many of the correctly predicted samples were incorrectly predicted by the models when we introduced the adversarial examples. Our result shows that deepseek6.7 and codellama are robust against this simple adversarial attack. However, llama showed the least robustness against this attack for 15 samples (out of 24 correct predictions); the model flipped the prediction label when the adversarial examples were introduced.    
\begin{table}[h!]
\captionsetup{position=top} 
\caption{Model Performance Comparison on Correct Prediction and Misclassification using Adversarial Samples}
\centering
\small  
\begin{tabular}{|l|c|c|}
\hline
\textbf{Model} & \textbf{Correct Predictions} & \textbf{Adversarial Misclassify} \\
\hline
deepseek6.7 & 25/50 & 1/25 \\
llama & 24/50 & 15/24 \\
codellama & 22/50 & 2/22 \\
\hline
\end{tabular}
\label{tab:misclassification_comparison}
\end{table}

\subsection{Case Studies}
\subsubsection{Google Translator vs LLM Translators on Evaluation}
We assume that the hate speech prediction results are not up to the mark because of the quality of the Google Translator \cite{googlecloudtranslation}. Google Translator API doesn't work well, preserving the sentiment of the actual language while translating to English. Hence, we decided to run a pilot study, if the LLMs can be used for the translation task effectively. We have prompted (Fig. 2) two LLMs for this task. First one is GPT 3.5-turbo \cite{openai_gpt_3_5_turbo}, and the other one is codellama-34b \cite{codellama_34b}. After the translation, we ran only the best-performing model (codellama-7b) from Table I for the hate speech detection task. For the translation part also, we select only one language, Bengali, with 200 comments. In TABLE IV, we have shown the comparative analysis of different translation techniques regarding different metrics. We expected better results when translating the language using LLMs. However, the table result doesn't show any improvement using LLMs over Google Translate. Our pre-trained LLMs translation capabilities are not there yet. We could have also tried different prompting strategies on the pre-trained models. However, that is beyond our current study.
\begin{table}[ht]
\captionsetup{position=top}
\caption{Performance Metrics for Different Translation Techniques on Codellama-7b Model}
\centering
\footnotesize 
\begin{tabular}{|l|c|c|c|c|}
\hline
\textbf{Translator} & \textbf{Accuracy} & \textbf{Precision} & \textbf{Recall} & \textbf{F1 Score} \\
\hline
Google Translator & 33.00 & 38.65 & 66.00 & 47.90 \\
GPT-3.5           & 31.00 & 38.27 & 62.00 & 47.33 \\
Codellama (34b)   & 31.50 & 39.75 & 66.00 & 49.62 \\
\hline
\end{tabular}
\label{tab:translation-techniques}
\end{table}

\subsubsection{Adversarial Misclassification Example}
As, we have mentioned in TABLE III that llama is working poorly on our synthetic adversarial dataset. Here we will be presenting a motivating example. 

\begin{figure}[h!]
\begin{flushleft}
\begin{tcolorbox}[width=\linewidth, colback=green!5, colframe=green!80, title=Original Comment]
"RT @psjohnson23: The problem with retard liberals is they think Supply \& Demand means ``Supply me with everything I Demand.'' \#VotingMatters"\\
\textbf{Label: Hate Speech} \\
\textbf{Llama Prediction: Hate Speech}
\end{tcolorbox}
\end{flushleft}
\caption{Example of an Original Comment Detected as Hate Speech}
\end{figure}

\begin{figure}[h!]
\begin{flushleft}
\begin{tcolorbox}[width=\linewidth, colback=red!5, colframe=red!80, title=Adversarial Comment]
"RT @psjohnson23: The issue with clueless thinkers is they assume ``Supply me with whatever I want'' is the rule. \#VotingMatters" \\
\textbf{Label: Hate Speech} \\
\textbf{Llama Prediction: Non Hate Speech}
\end{tcolorbox}
\end{flushleft}
\caption{Example of an Adversarial Comment Misclassified as Non Hate Speech}
\end{figure}

In Fig. 6, we present a comment from the original dataset labeled as hate speech and Llama also predicted the comment as hate speech. After generating adversarial comments, the prediction of Llama has been flipped, as shown in Fig. 7.

\section{Discussion}
The evaluation result section demonstrates the capabilities and limitations of pre-trained LLMs in Hate Speech Detection. Our result reveals significant differences in binary hate speech detection on different LLMs. Codellama-7b achieved better results than Llama-2 and deepseekcoder in all metrics, especially in recall accuracy. However, the accuracy of codellama-7b was still suboptimal compared to its higher recall, suggesting the tendency of the model to be biased towards the hate speech class. \\
The limitation of codellama-7b is also evident in the case of correct geographic prediction. Although codellama-7b achieved the highest accuracy for predicting hate speech, the percentage of predicting correct geography is notably poor. Deekseekcoder achieved the highest number of correct geographic predictions, and Llama-2 achieved the best percentage in predicting the correct location, indicating promising aspects of further targeted fine-tuning.\\
We also analyzed the adversarial robustness of the three LLM models. Our study shows Llama-2 is highly susceptible to adversarial attacks, flipping more than half correct predictions during the attack. Codellama-7b and deepseekcoder demonstrate substantial robustness against our adversarial attack. \\
We conducted a pilot study to find out the effectiveness of LLM as a translator vs Google Translator. Our analysis shows using LLM to translate the original language doesn't help to improve the hate speech prediction performance. However, a detailed study on this may be required to determine the actual reason. 

\section{Threats to Validity}
Our original dataset had a total of 10k samples. Out of those, we selected 1k samples for evaluation randomly. We haven't used any other systematic filtering or heuristics to choose the "better" samples for evaluation. We didn't find many LLMs working on Multilingual data, so we had to translate the original hate speech comment into English. So, our work is heavily dependent on the quality of the translation. Our synthetic adversarial dataset contains only 50 samples, which is small set of evaluation. Additionally, While adversarial examples were used to test robustness, these synthetic samples might not reflect real-world adversarial behaviors, potentially limiting the applicability of the findings to real-world scenarios. Lastly, our work is fully evaluated on pre-trained models, which might not be good enough for this specialized task. 

\section{Conclusion and Future Work}
In this study, we focused on analyzing the capabilities of the LLMs on multilingual hate speech detection and finding out the geographic context of the hate speech. We translated the original language comment to English and demonstrated the feasibility of leveraging LLMs in a multilingual hate speech dataset. We have curated 10k samples across five languages and evaluated three LLMs on 1k randomly selected samples. Our approach heavily relies on the quality of the translation, which is a major dependency of our study. Despite the challenges, our work will contribute to researching advanced multilingual hate speech detection and translation quality. \\
In the future, we would like to work on targeted fine-tuned LLMs. The 10k samples we curated will not be enough to fine-tune LLMs, so we must curate more labeled data from various public sources. It would also be great to see how the new LLMs will perform on our multilingual dataset; we couldn't run more than three models for this study due to resource constraints. \\

\section*{DATA AVAILABILITY}
Our dataset and codes are available \href{https://github.com/Monoshi-tonmoy/Evaluating-LLMs-on-Hate-Speech-Detection}{here}.

\section*{Acknowledgment}
We want to express our gratitude to Professor Amit Kumar Sikder for his valuable feedback and guidance on this project.


\end{document}